\documentclass[letterpaper, 10 pt, conference]{ieeeconf}  

\IEEEoverridecommandlockouts                              

\title{\LARGE \bf
Social Robots As Companions for Lonely Hearts: The Role of Anthropomorphism and Robot Appearance
}

\author{Yoonwon Jung$^{1}$ and Sowon Hahn$^{2}$
\thanks{*This work was not supported by any organization}
\thanks{$^{1}$Yoonwon Jung is with the Department of Psychology, Seoul National University,
        Seoul 08826, South Korea
        {\tt\small ywjung@snu.ac.kr}}%
\thanks{$^{2}$Sowon Hahn is a Faculty of the Department of Psychology, Seoul National University,
        Seoul 08826, South Korea
        {\tt\small swhahn@snu.ac.kr}}%
}

\usepackage{tabularray, tabularx}
\usepackage{booktabs}
\usepackage{multirow}
\usepackage{fancyhdr}
\usepackage{graphicx}
\usepackage{caption}
\usepackage{subcaption}
\usepackage[indentfirst=false,rightmargin=0pt]{quoting}
\usepackage{makecell}

\begin{document}

\maketitle
\thispagestyle{empty}
\pagestyle{empty}

\begin{abstract}

Loneliness is a distressing personal experience and a growing social issue. Social robots could alleviate the pain of loneliness, particularly for those who lack in-person interaction. This paper investigated how the effect of loneliness on the anthropomorphism of social robots differs by robot appearance, and how it influences purchase intention. Participants viewed a video of one of the three robots (machine-like, animal-like, and human-like) moving and interacting with a human counterpart. Bootstrapped multiple regression results revealed that although the unique effect of animal-likeness on anthropomorphism compared to human-likeness was higher, lonely individuals' tendency to anthropomorphize the animal-like robot was lower than that of the human-like robot. This moderating effect remained significant after covariates were included. Bootstrapped mediation analysis showed that anthropomorphism had both a positive direct effect on purchase intent and a positive indirect effect mediated by likability. Our results suggest that lonely individuals' tendency of anthropomorphizing social robots should not be summarized into one unified inclination. Moreover, by extending the effect of loneliness on anthropomorphism to likability and purchase intent, this current study explored the potential of social robots to be adopted as companions of lonely individuals in their real life. Lastly, we discuss the practical implications of the current study for designing social robots.
\end{abstract}

\section{INTRODUCTION}
The need to belong is fundamental to humans \cite{RN169}. When this need for social connection is unmet at the desired magnitude, individuals experience loneliness \cite{PP1981}. Loneliness is linked to increased depressive symptoms and cognitive decline, and is also associated with increased risks of getting cardiovascular diseases and compounded immune systems \cite{HC2010}. These grave mental and physical declines lead to higher medical and general practitioner costs \cite{Betal2021,Meetal2021}.

In situations where in-person contact is limited, such as in solitary living arrangements, social robots could be a potential source of replenishing the sense of connection \cite{B2003}. Indeed, empirical evidence demonstrated that social robots can effectively alleviate feelings of loneliness \cite{Betal2008,Retal2013}. Therefore, this paper explored the potential of social robots as real-life companions of lonely individuals by investigating the factors that contribute to positive human-robot interaction and enhance adoption intention.

Among the factors that shape lonely individuals' interaction with social robots, anthropomorphism plays an important role. Anthropomorphism enhances the human counterparts' likability and trust in social robots \cite{Eetal2010,RN116}. This favorable attitude, caused by increased anthropomorphism, leads to customers' increased likelihood of purchasing technology-driven products (i.e. chatbots). Previous research has identified a positive relationship between anthropomorphism and customers' likelihood of purchasing technology-driven products (i.e. web design, chatbot), with likability or enjoyment acting as a mediating factor \cite{RN181,RN182,RN207}. 

Lonely individuals generally show a heightened tendency to anthropomorphize non-human objects and entities \cite{RN99,RN100,RN131,RN112,ER2013}. However, empirical evidence on the anthropomorphic inclination of lonely individuals toward social robots is inconsistent. One suggested that loneliness increases the inclination to anthropomorphize social robots \cite{ER2013}, while another study reported that loneliness decreases this tendency \cite{RN122}. Given that lonely individuals are prone to perceiving social cues as potential threats \cite{HC2010,CH2009}, robot appearance could be an important factor that shapes lonely individuals' anthropomorphic tendency toward social robots. Yet, no research has examined the differential effects of robot appearance on this anthropomorphic tendency. 

In light of the aforementioned reasons, this paper aims to shed light on the intricate interplay between loneliness and robot appearance on the perception of social robots as human-like entities. Furthermore, we seek to uncover how individuals' inclination to anthropomorphize social robots influences their intention to purchase social robots. 

\section{Related Work}

\subsection{Anthropomorphism}

Humans anthropomorphize non-human objects or agents by attributing human-like characteristics (i.e. motivations, intentions, emotions) to them \cite{RN98}. A group of scholars defined two dimensions of humanness traits as human nature (HN) and uniquely human nature (UHN) \cite{RN101,RN107}. HN encompasses traits shared by humans and animals, while UHN represents characteristics exclusive to humans. Similarly, another group of scholars suggested two dimensions of mind perception as agency and experience \cite{RN88}. Experience represents bodily sensations and feelings, whereas agency refers to the capacity for intentional action. There is a notable overlap between the two concepts. Agency and experience align with UHN and HN, respectively \cite{RN88}. 

Empirical studies have shown that individuals perceive typical humans as possessing high levels of both agency and experience, mammals as low in agency but high in experience, and robots as high in agency but low in experience \cite{RN88,RN134}. This suggests that when people anthropomorphize non-human objects or agents, they attribute the dimensions that they perceive to be lacking in the nature of these objects or agents. Indeed, it has been argued that attributing perceived experience to robots would help reduce their perceived machine-likeness, and imbue them with a more human-like bearing \cite{RN122}. Therefore, this paper adopts perceived experience as a dimension of human likeness that people imbue on social robots when anthropomorphizing them.

\subsection{Relationship between loneliness and anthropomorphism}

Previous research proposed a three-factor theory of anthropomorphism \cite{RN99, RN98}. One of the factors, sociality motivation, represents the desire for social connection and attachment, which suggests that lonely individuals compensate for unmet social needs by anthropomorphizing non-human entities \cite{RN100}. Indeed, empirical evidence demonstrated that lonely individuals showed heightened inclination to perceive human-like attributes in various entities (i.e., animals, technical gadgets, robotic heads) \cite{RN99,RN100,RN131,RN112,ER2013}.

On the other hand, a recent study reported that trait loneliness lowered the tendency to anthropomorphize and accept a human-like social robot \cite{RN122}. The authors suggested that dispositionally lonely individuals, who are more prone to interpret their social surroundings as threatening \cite{HC2010,CH2009}, may have perceived the robot as unsettling. 

Categorizing the studies based on the types of non-human entities used, a discernible pattern emerges in these inconsistencies. Lonely individuals tend to exhibit higher levels of anthropomorphism towards simple technical devices and animals \cite{RN99,RN100,RN131,RN112}. However, research involving robots reveals mixed results, reporting both positive and negative associations between loneliness and anthropomorphism \cite{ER2013,RN122}. These disparities indicate potential variations in the anthropomorphic tendencies of lonely individuals specifically towards social robots, distinguishing them from other non-human entities. However, further investigation is needed to validate this, as the existing body of research on this topic that used sophisticated robots remains limited.

\subsection{The effect of robot appearance on lonely individuals' anthropomorphic tendency toward social robots}

The research on anthropomorphizing social robots has mostly used either machine-like robots or human-like robots. Considering that human-like robots have been tested as a kind of gold standard for robot anthropomorphism, this may sound trivial. However, previous research on more general anthropomorphic tendencies of humans suggests that humans perceive humanness from not only human-looking entities but also from animals (e.g. pets) and machine-like objects (i.e. clocks, cars) \cite{RN99,RN100,RN131,RN112,ER2013}. Thus, when integrating general research on anthropomorphism into robotics research, exploring the varying degree of anthropomorphism across different robot appearances is needed.

Therefore, we used three robots with different appearances (human-like, machine-like, and animal-like) to investigate whether the tendency of lonely individuals to anthropomorphize social robots differs by robot appearance. Although the robot used in \cite{RN122} had a head with two eyes and a torso, the robot was generally more machine-like compared to that in \cite{ER2013}. Thus, we hypothesized that machine-likeness would decrease lonely individuals’ tendency to anthropomorphize. Moreover, since human-like characteristics are more likely than animal-like characteristics to evoke anthropomorphism, we predicted that lonely individuals would anthropomorphize animal-like robots less than human-like ones.

\begin{quoting}[leftmargin=\parindent]
\textbf{H1}: Lonely individuals' anthropomorphic tendencies will differ by social robots' appearance.  \\
\textbf{H1-1}: Lonely individuals will show a lower tendency to anthropomorphize a machine-like robot than a human-like robot.\\
\textbf{H1-2}: Lonely individuals will show a lower tendency to anthropomorphize an animal-like robot than a human-like robot.\\
\end{quoting}

\subsection{The effect of anthropomorphism on the likability and purchase intent of social robots}

To facilitate the real-world adoption of social robots by lonely individuals, increasing the acceptance motivation is pivotal. Purchase intention, in particular, signifies the willingness to invest financially to utilize the robots in the users' everyday life. Thus, investigating how anthropomorphizing social robots affect purchase intention will foreground the practical aspect of investigating the relationship between loneliness and anthropomorphism.

Attributing human traits to social robots increases likability and trust among human counterparts \cite{Eetal2010,RN116}. Previous research has demonstrated that anthropomorphism increases purchase intent, and perceived enjoyment or likability acts as a mediating factor \cite{RN181,RN182,RN207}. However, the association was tested on non-robotic products, web design, and disembodied chatbots. Therefore, additional investigation is needed to examine whether the relationship between anthropomorphism and purchase intent, along with the mediation effect of likability, extends to embodied social robots.

Moreover, whether lonely individuals' heightened anthropomorphism leads to higher purchase intent remains unexplored. Although one study investigated the moderating role of loneliness in the association between anthropomorphism and consumer attitudes \cite{RN207}, the anthropomorphized objects were brands and advertisements. Therefore, exploring the behavioral consequences of lonely individuals' anthropomorphic inclination in relation to social robots is crucial. We predicted that if the effect of loneliness on anthropomorphizing social robots varies with robot appearance, such differences would also be reflected in  purchase intent.

\begin{quoting}[leftmargin=\parindent]
\textbf{H2}: Higher anthropomorphism will lead to increased purchase intent via increased likability of social robots.  \\
\textbf{H3}: Differences in lonely individuals' anthropomorphic tendencies by robot appearance will predict differences in the purchase intent of social robots.\\
\end{quoting}

\section{Study Design}

\subsection{Participants}

Participants for this study were recruited from multiple sources, including the participant recruitment system of Seoul National University and online student communities of both Seoul National University and Korea University. This study was approved by Seoul National University Institutional Review Board.

From January to May of 2022, participants responded online through the survey built using Qualtrics. Therefore, we conducted extensive quality control to strictly judge the quality of responses. Participants who failed one of the two attention-check items (i.e. “For this question, please check ‘strongly agree’”) or completed the survey too quickly were not considered to be fully engaged in the study \cite{RN131}. Those who failed the attention check were unable to proceed with the survey as it terminated upon an incorrect response. Moreover, we measured the average time needed to fully watch the video and attentively respond to the questionnaires from a pilot study, and concluded that a proper response needs at least 320 seconds. Consequently, we excluded the responses that took less than 320 seconds to complete.

\subsection{Materials}

\subsubsection{Robots}
\begin{figure}[t]
  \centering
  \includegraphics[width=0.9\linewidth]{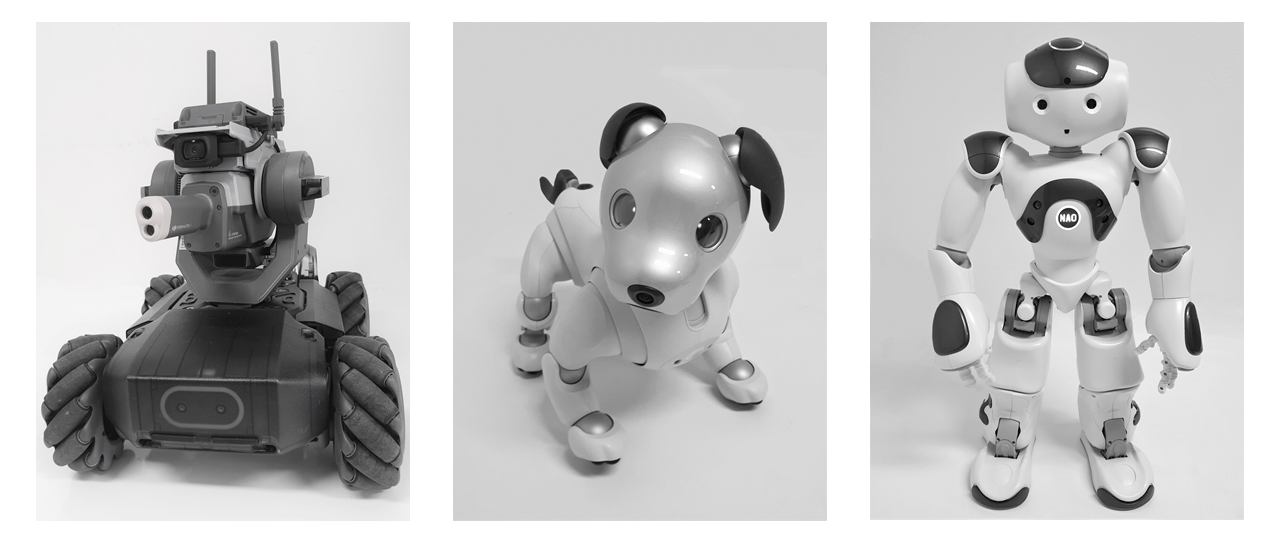}
  \caption{Robots of three robot appearance conditions}
  \label{fig.1}
\end{figure}

We used DJI Robomaster S1, Softbank Robotics' NAO, and SONY's AIBO formachine-like, human-like, and animal-like robots, respectively (Fig. \ref{fig.1}).

\subsubsection{Human-robot interaction videos}
\begin{figure}[t]
  \centering
    \begin{subfigure}[t]{0.7\linewidth}
         \centering
         \includegraphics[width=\textwidth]{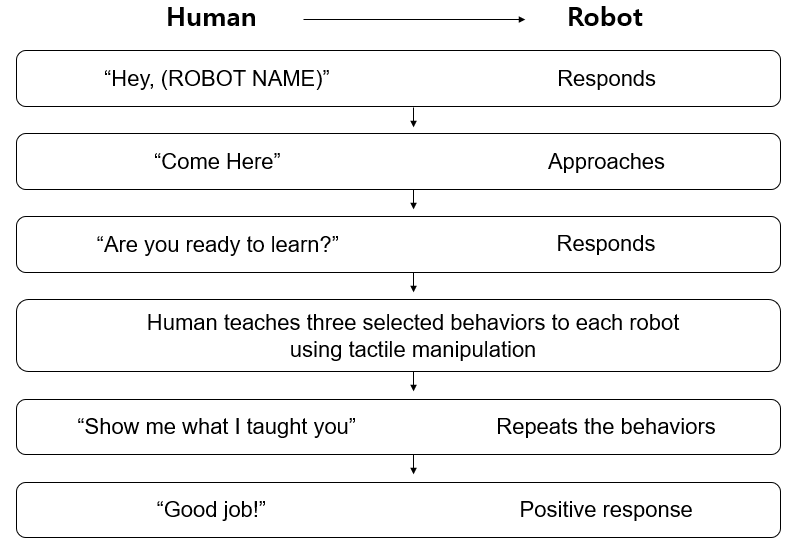}
         \caption{Flow chart of human-robot interaction}
         \label{fig2a}
    \end{subfigure}
    \begin{subfigure}[t]{0.2\columnwidth}
         \centering
         \includegraphics[width=\textwidth]{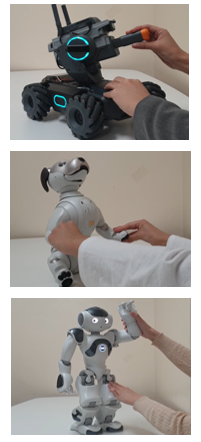}
         \caption{Tactile \newline manipulation}
         \label{fig2b}
    \end{subfigure}
  \caption{Human-robot interaction scenario and examples}
  \label{fig.2}
\end{figure}

The interactions between robots and human counterparts were recorded as videos. The robots engaged with their counterparts following the same interaction scenario across all three conditions (Fig. \ref{fig.2}). Robots responded 1.8 seconds after the human voice, and the lengths of videos were around 90 seconds.

Three robots reflected different levels of lifelikeness during the interaction. To the human counterparts' speech, the machine-like robot responded using LED light or a basic beep sound. It also rotated its upper body or moved upfront to respond. The animal-like robot produced dog-like sounds and imitated the dog behaviors using ear flop, head movement, mouth opening, tail wagging, sitting, and front paw lowering. The human-like robot answered using spoken words and conveyed nonverbal cues by moving its head or limbs. Moreover, robots showed distinct positive responses to their human counterparts' praise. The machine-like robot blinked its light, produced a beeping sound twice, and quickly rotated left and right. The animal-like robot barked and wagged its tail, and then lowered its head and front paw. The human-like robot showed a proud gesture and said "Thank you!".

We used Choreographe and Robomaster desktop application to program NAO and Robomaster S1, respectively. The basic functions of AIBO were used without programming. For the videos, We first filmed NAO and Robomaster executing the pre-programmed behaviors without human commands and added Korean human speech afterward. For AIBO, we filmed the interaction using Japanese commands and replaced the commands with Korean human speech.

\subsubsection{Self-reported measures}

To assess loneliness, we used the Korean-validated version of the UCLA loneliness scale version 3 \cite{RN83}. We used 4-point Likert scales, ranging from ‘never’ to ‘often’. The internal reliability was 0.94. To measure anthropomorphism, we used scales based on the 2-dimensional model of mind perception \cite{RN88}. Following studies suggesting the reduced version of the scales consisting of items with more distinguishable factor loadings \cite{RN161,RN115}, we used 6 items for each dimension. Seven-point scales ranging from 1 to 7 were used. The internal reliability was 0.85 for perceived experience and 0.75 for perceived agency.

For likability, we used 3-item differential scales \cite{RN90}. Seven-point scales ranging from -3 to 3 were used, with -3 and 3 representing antipole adjectives (e.g. 'unlikable - likable'). The internal reliability was 0.9. For purchase intent, we used a single-item measure. Participants used 7-point scales ranging from 1 to 7. We also measured variables that could be covariates: age, gender, household income, and prior knowledge of robots \cite{RN162,RN160,RN195}.

\begin{table}[b]
  \renewcommand{\arraystretch}{1.1}
  \centering
  \captionsetup{width=\linewidth}
  \caption{Means and standard deviations of variables by robot appearance}
  \label{tab:1}
  \setlength\tabcolsep{2.5pt} 
  \resizebox{\columnwidth}{!}{
    \begin{tabular}{clccccc}
    \toprule\tabularnewline[-1em]
     &
       &
      Loneliness &
      {\shortstack{Perceived\\ Experience}} &
      \multicolumn{1}{l}{\shortstack{Perceived\\ Agency}} &
      Likability &
      {\shortstack{Purchase\\ Intent}} \\[+0.2em] \midrule
    \multirow{2}{*}{\shortstack{Animal-like\\(\emph{N}=46)}}  & \emph{M}                            & 2.330 & 2.122 & 3.307 & 1.082  & 3.330 \\[-0.2em]
                                                                                    & \multicolumn{1}{c}{\textit{SD}} & 0.586 & 1.163 & 1.240 & 1.140  & 1.398 \\
    \multirow{2}{*}{\shortstack{Human-like\\(\emph{N}=45)}}   & \emph{M}                            & 2.284 & 2.063 & 4.074 & 0.644  & 2.733 \\[-0.2em]
                                                                                    & \multicolumn{1}{c}{\textit{SD}} & 0.064 & 1.475 & 1.296 & 1.293  & 1.572 \\
    \multirow{2}{*}{\shortstack{Machine-like\\(\emph{N}=46)}} & \emph{M}                            & 2.265 & 1.475 & 2.872 & -0.723 & 2.489 \\[-0.2em]
                                                                                    & \multicolumn{1}{c}{\textit{SD}} & 0.549 & 0.650 & 1.403 & 1.254  & 1.666 \\ \midrule
    \multirow{2}{*}{\shortstack{Total\\(\emph{N}=137)}}                                                          & \emph{M}                            & 2.293 & 1.881 & 3.410 & 0.309  & 2.847 \\[-0.2em]
                                                                                    & \multicolumn{1}{c}{\textit{SD}} & 0.576 & 1.001 & 1.399 & 1.443  & 1.581 \\ \bottomrule
\end{tabular}
}
\caption*{\raggedright\footnotesize Note: All variables were standardized.}
\end{table}

\subsection{Procedure}

Participants completed questionnaires measuring loneliness. Then, participants were randomly assigned to one of three robot appearance conditions, and watched a video of a robot interacting with a human counterpart. Lastly, participants filled out questionnaires regarding their perception of the robot, along with demographic variables and their prior knowledge and familiarity with the robot. Participants who completed the study received the promised reward.

\section{Results}
Of 185 total responses, 137 (${N}_{female}$=84, ${M}_{age}$=22.803, ${SD}_{age}$=3.788) were selected as final data. It consisted of 46 responses from the animal-like condition, 45 from the human-like condition, and 46 from the machine-like condition. See Tab. \ref{tab:1} for other descriptive statistics.

\begin{table}[b]
\renewcommand{\arraystretch}{1.05}
  \captionsetup{width=\linewidth}
    \caption{Bootstrapped regression results on robot appearance, loneliness, and their interactions as predictors for perceived experience}
    \label{tab:2}
      \begin{tabular}{lccc}
        \toprule
        &\emph{$\beta$}&\emph{SE}&95\% \emph{BCa CI}\\[-0.1em]
        \midrule
        Intercept&-0.101&0.147&[-0.341, 0.235]\\
        Robot1(machine-like=1)&-0.147&0.185&[-0.537, 0.200]\\
        Robot2(animal-like=1)&\textbf{0.470*}&0.221&[0.045, 0.896]\\
        Loneliness&0.197&0.161&[-0.053, 0.587]\\
        Robot1-loneliness interaction&-0.288&0.192&[-0.699, 0.060]\\
        Robot2-loneliness interaction&\textbf{-0.500*}&0.260&[-1.080, -0.056]\\ 
      \bottomrule
    \end{tabular}
    \caption*{\footnotesize Note: Robot1 and Robot2 are contrasts of robot appearance levels (Robot1; machine-like robot compared to human-like robot, Robot2;animal-like robot compared to human-like robot). All variables were standardized. Bootstrapped \emph{$R^2$} value was 0.122 (95\% \emph{BCa CI}= [0.025, 0.219]). \newline
    *= statistically significant at 95\% confidence level}
\end{table}

\begin{table}[b!]
\renewcommand{\arraystretch}{1.05}
  \captionsetup{width=\linewidth}
    \caption{Bootstrapped regression results on robot appearance, loneliness, and their interactions as predictors for perceived experience with covariates}
    \label{tab:3}
      \begin{tabular}{lccc}
        \toprule
        &\emph{$\beta$}&\emph{SE}&95\% \emph{BCa CI}\\[-0.1em]
        \midrule
        Intercept&-0.060&0.166&[-0.361, 0.287]\\
        Age&0.054&0.109&[-0.152, 0.288]\\
        Gender(male=1)&-0.169&0.177&[-0.532, 0.176]\\
        Income&-0.033&0.086&[-0.215, 0.133]\\
        Prior knowledge on robot&\textbf{0.175*}&0.092&[0.013, 0.375]\\
        Robot1(machine-like=1)&-0.095&0.166&[-0.487, 0.291]\\
        Robot2(animal-like=1)&\textbf{0.485*}&-0.220&[0.062, 0.938]\\
        Loneliness&0.167&0.160&[-0.130, 0.521]\\
        Robot1-loneliness interaction&-0.253&0.199&[-0.653, 0.127]\\
        Robot2-loneliness interaction&\textbf{-0.496*}&0.249&[-1.070, -0.060]\\
      \bottomrule
    \end{tabular}
    \caption*{\footnotesize Note: Robot1 and Robot2 are contrasts of robot appearance levels (Robot1; machine-like robot compared to human-like robot, Robot2;animal-like robot compared to human-like robot). All variables were standardized. Bootstrapped \emph{$R^2$} value was 0.153 (95\% \emph{BCa CI}= [0.054, 0.242]). \newline
    *= statistically significant at 95\% confidence level}
\end{table}

\begin{figure}[b!]
  \centering
  \captionsetup{width=.\linewidth}
  \includegraphics[width=0.92\linewidth]{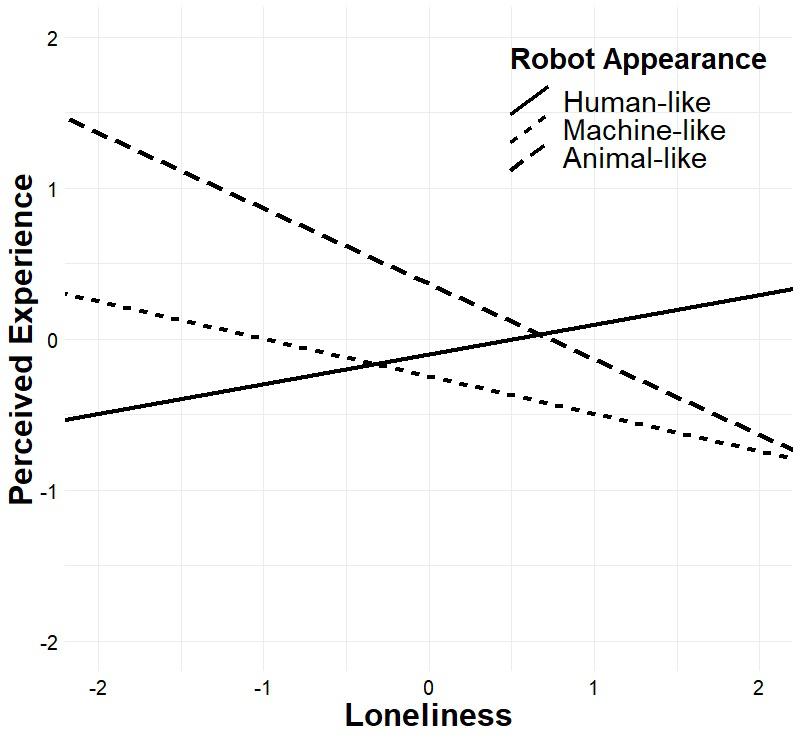}
  \caption{The effect of loneliness on perceived experience by robot appearance conditions}
  \label{fig.3}
  \caption*{\raggedright\footnotesize Note: All variables were standardized.}
\end{figure}

We conducted bootstrapped multiple regression to verify the impact of loneliness on robot anthropomorphism and the moderation effect of robot appearances (Tab. \ref{tab:2}-\ref{tab:3} and Fig. \ref{fig.3}). \texttt{CAR} and \texttt{boot} packages were used for bootstrapping regression coefficients, R-squared values, and their BCa 95\% confidence intervals \cite{RN186,RN185} in R. Robot appearances were coded as two dummy variables; robot1 (machine-like robot compared to human-like robot) and robot2 (animal-like robot compared to human-like robot). We used 2000 iterations with seeds set as 123 for reproducibility. 

The unique effect of loneliness on perceived agency was not significant (\emph{$\beta$}=.115, \emph{SE}=.180, \emph{CI}=[-.250, .476]), nor the moderation effects of robot appearance (robot1-loneliness interaction: \emph{$\beta$}=-.170, \emph{SE}=.227, \emph{CI}=[-.599, .291]; robot2-loneliness interaction: \emph{$\beta$}=-.303, \emph{SE}=.231, \emph{CI}=[-.758, .168]).

\begin{table*}[ht!]
\renewcommand{\arraystretch}{1.2}
\centering
\captionsetup{width=0.93\textwidth}
\caption{Summary of the Results of the PROCESS model7}
\label{tab:5}
      \begin{tabular}{c c c c c c}
        \toprule
        \multicolumn{2}{c}{\shortstack{Moderated mediation test \\ \emph{$\beta$(SE)}, 95\% \emph{BCa CI}}} & \multicolumn{3}{c}{\shortstack{Conditional Indirect Effect \\ \emph{$\beta$(SE)}, 95\% \emph{BCa CI}}} & \multirow{2}{*}{\shortstack{Direct Effect \\ \emph{$\beta$(SE)}, 95\% \emph{BCa CI}}} \\ \cline{1-5}
 Robot1  &Robot2  &Human-like  &Machine-like  &Animal-like  & \\[-0.2em]
        \midrule
        \shortstack{-0.099(0.080), \\ CI=[-0.277, 0.035]}  &\shortstack{\textbf{-0.194*}(0.108), \\ CI=[-0.445, -0.016]}&\shortstack{0.065(0.064), \\  CI=[-0.045, 0.203]}&\shortstack{-0.034(0.045), \\ CI=[-0.124, 0.057]}&\shortstack{-0.129(0.084), \\  CI=[-0.308, 0.015]}&\shortstack{0.047(0.083), \\ CI=[-0.117, 0.210]}\\[-0.1em]
        \bottomrule
    \end{tabular}
    \caption*{\footnotesize Note: Prior knowledge of robots, age, gender, and income were the covariates of the models. All variables were standardized. \newline
    *= statistically significant at 95\% confidence level}
\end{table*}

The unique effect of loneliness on perceived experience was not significant (\emph{$\beta$}=.197, \emph{SE}=.161, \emph{CI}=[-.053, .587]). The interaction effect between robot1 and loneliness was not significant (\emph{$\beta$}=-.288, \emph{SE}=.192, \emph{CI}=[-.699, .060]). However, the interaction term between robot2 and loneliness was significant and negative (\emph{$\beta$}=-.500, \emph{SE}=.260, \emph{CI}=[-1.080, -.056]). This interaction effect persisted when covariates were introduced (\emph{$\beta$}=-.496, \emph{SE}=.249, \emph{CI}=[-1.070, -.060]). This interaction effect is in contrast with the positive unique effect of robot2 (\emph{$\beta$}=.470, \emph{SE}=.221, \emph{CI}=[-.045, .896]).

\begin{figure}[t]
  \centering
  \includegraphics[width=0.83\linewidth]{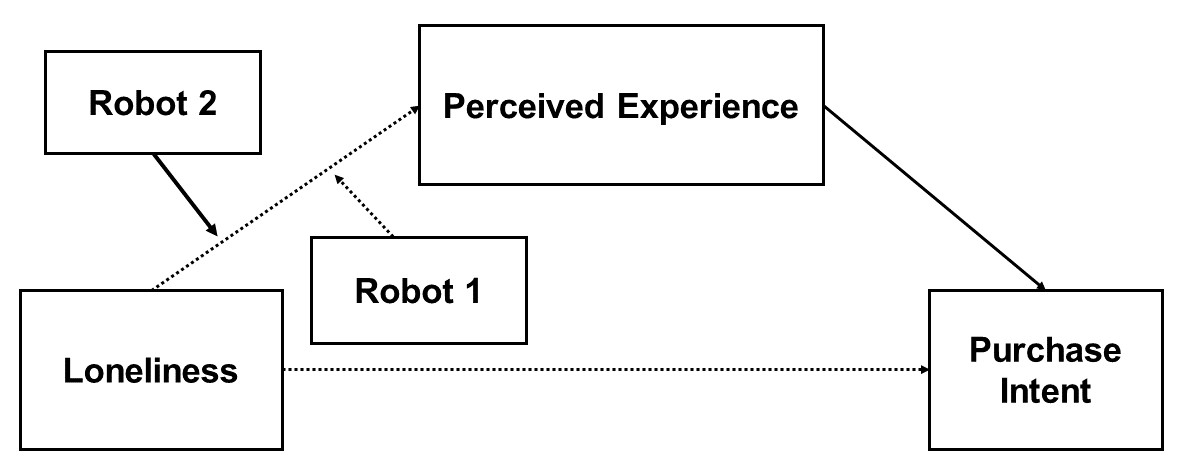}
  \captionsetup{width=.48\textwidth}
  \caption{Overview of the results of the \texttt{PROCESS} model7}
  \label{fig.4}
  \caption*{\footnotesize Note: Robot1 and Robot2 are contrasts of robot appearance levels (Robot1; machine-like robot compared to human-like robot, Robot2;animal-like robot compared to human-like robot.) Solid lines indicate significant relationships, while dotted lines indicate non-significant relationships. Prior knowledge of robots, age, gender, and income were the covariates.}
\end{figure}

Also, we tested the indirect and direct effect of anthropomorphism on purchase intent using \texttt{PROCESS macro} in R \cite{RN66}. The mediator was likability. We included prior knowledge of robots, gender, and income as covariates, which were the significant factors in Tab. \ref{tab:3}. We used 1000 iterations with seeds set as 100. Perceived agency had a significant positive indirect effect on purchase intent (\emph{$\beta$}=.225, \emph{SE}=.545, \emph{CI}=[.119, .332]) but the direct effect was not significant (\emph{$\beta$}=.069, \emph{SE}=.075, \emph{CI}=[-.081, .218]). Perceived experience had both a significant positive indirect effect (\emph{$\beta$}=.135, \emph{SE}=.046, \emph{CI}=[.043, .233]) and direct effect (\emph{$\beta$}=.250, \emph{SE}=.069, \emph{CI}=[.115, .386]) on purchase intent. 

Lastly, we tested the final model using \texttt{PROCESS macro} in R \cite{RN66}. We used 1000 iterations with seeds set as 100. Prior knowledge of robots, age, gender, and income were the covariates. Since the direct effect of anthropomorphism on purchase intent was significant, likability was excluded from the model for simplicity. We included perceived experience, the indicator of social robot anthropomorphism, as a mediator of the model. The moderated mediation was significant when the animal-like robot was compared to the human-like robot. See Tab. \ref{tab:5} and Fig. \ref{fig.4} for the detailed results.

\section{Discussion}
This study examined the effect of loneliness on anthropomorphizing social robots and the moderating effect of robot appearance. We also investigated how lonely individuals' anthropomorphic tendency predicts purchase intent. 

Results revealed that lonely individuals' anthropomorphic tendency was lower in the animal-like robot than in the human-like one. This effect remained significant after covariates were introduced. Thus, H1 and H1-2 were supported.

The rejection of H1-1 implies that the machine-like robot did not significantly differ from the human-like robot in inducing lonely individuals to anthropomorphize. However, although not statistically significant, the regression coefficients show that while loneliness and anthropomorphism were positively associated in the human-like robot condition, the association was negative in the machine-like robot condition. This implies that the degree of machine-likeness could partly explain the inconsistency in lonely individuals’ anthropomorphizing tendency toward social robots \cite{ER2013,RN122}.

Our findings supporting H1-2 suggests that lonely individuals are less inclined to anthropomorphize animal-like robots (e.g., AIBO) compared to human-like robots. This is in contrast with our other finding that participants generally anthropomorphized the animal-like robot more compared to the human-like robot. One possible explanation for this is that AIBO displaying a dog-like resemblance in an adorable way increased the general anthropomorphic tendency, but not for lonely individuals. AIBO was designed as a ‘cute companion dog’ by 'showing moves and gestures in adorable patterns' \cite{RN205}. It mimics not only the visual appearance but also the movements and sounds of a real dog seeking affection. For lonely people who are hypervigilant to social threats \cite{HC2010,CH2009}, which makes them easier to perceive non-harmful social cues as threatening, such features could have been perceived as unsettling or potentially threatening. 

Furthermore, anthropomorphism had a positive direct effect and an indirect effect via likability on purchase intent. Moreover, the mediated moderation index was significant when the animal-like robot was compared to the human-like robot. Specifically, the mediating effect significantly decreased in animal-like robots compared to human-like robots, reflecting the H1-2 being supported. Thus, our findings support H2 and H3. This indicates that if a social robot induces higher anthropomorphism from lonely individuals, this leads lonely individuals to like the robot more, and see more value in the acquisition of the robot \cite{RN207}. 


This study makes several theoretical contributions. Above all, this paper is the first to use animal-like and machine-like robots, along with human-like robots, to compare lonely individuals’ anthropomorphic tendencies across different robot appearances. The results indicate that the inclination of lonely individuals to anthropomorphize social robots should be evaluated in relation to the specific characteristics of robots, rather than unifying it as one general inclination. Furthermore, we extend lonely individuals' anthropomorphic tendencies toward social robots to their adoption intent as consumers. By examining the possibilities of social robots as companions for lonely individuals in their real life, our results strengthen the importance of studying lonely individuals' anthropomorphic tendencies toward social robots. Lastly, this paper carries implications for future loneliness intervention research. Loneliness intervention research using robots mainly focused on animal-like robots (e.g. \cite{Betal2008,Retal2013}). Future studies testing the loneliness-mitigating effect of social robots should also test the effect of humanoid robots, which were shown to induce more anthropomorphism and likability toward social robots from lonely individuals.

This study also provides insights for designing social robots for lonely individuals. Robots should evoke anthropomorphism only to match the degree of realism of their appearance. If it fails to do so, lonely individuals could perceive social robots as threatening \cite{HC2010,CH2009}. We suggest social robots demonstrate suitable levels of movement and interaction skills to ensure they maintain congruity with their appearance and avoid excessive mimicry of lifeforms.

Despite the valuable findings and implications of this study, they should be regarded with caution due to some limitations.  Firstly, data collection was conducted online due to COVID-19 safety concerns. Although we conducted extensive quality control measures, this could have potentially affected participant engagement. Additionally, the absence of on-site viewing of the interaction raises the need for future studies to replicate our study using in-person experiments that involve participants physically interacting with the robot. Lastly, the relationships could be tested in an experimental field study, increasing the results’ ecological validity.

\bibliographystyle{IEEEtran}
\bibliography{IEEEabrv,mybib}

\end{document}